\documentclass[preprint,12pt]{elsarticle}

\usepackage[top=1in, bottom=1in, left=1in, right=1in]{geometry}

\usepackage[english]{babel}
\usepackage[utf8]{inputenc}
\usepackage{amsmath,amssymb}
\usepackage{parskip}
\usepackage{graphicx}
\usepackage{subfig}
\usepackage{pifont}
\usepackage{diagbox}
\usepackage{booktabs,tabularx}
\usepackage[title]{appendix}
\usepackage{pdfpages}

\usepackage[colorlinks=true, bookmarksopen,
            pdfsubject={algorithms},
            linkcolor={blue},
            anchorcolor={black},
            citecolor={black},
            filecolor={magenta},
            menucolor={black},
            plainpages=false,pdfpagelabels,
            urlcolor={blue}]{hyperref}
            
\newcommand{\bz}{\boldsymbol{z}}
\newcommand{\bu}{\boldsymbol{u}}
\newcommand{\bp}{\boldsymbol{p}}
\newcommand{\bq}{\boldsymbol{q}}
\newcommand{\bb}{\boldsymbol{b}}
\newcommand{\bx}{\boldsymbol{x}}
\newcommand{\by}{\boldsymbol{y}}
\newcommand{\bw}{\boldsymbol{w}}

\newcommand{\bR}{\mathbb{R}}
\newcommand{\mD}{\mathcal{D}}
\newcommand{\mI}{\mathcal{I}}

\newcommand{\cmark}{\ding{51}}%
\newcommand{\xmark}{\ding{55}}%

\newtheorem{rem}{Remark}

\begin{document}

\begin{frontmatter}
\title{Improving the Expressive Power of Deep Neural Networks through Integral Activation Transform}
\author[a]{Zezhong Zhang}
\author[b]{Feng Bao}
\author[a]{Guannan Zhang\vspace{-0.3cm}}

\address[a]{Computer Science and Mathematics Division, Oak Ridge National Laboratory, Oak Ridge, TN 37831.\vspace{-0.2cm}}
\address[b]{Department of Mathematics, Florida State University, Tallahassee, FL 32306.}

\begin{abstract}
The impressive expressive power of deep neural networks (DNNs) underlies their widespread applicability. However, while the theoretical capacity of deep architectures is high, the practical expressive power achieved through successful training often falls short. Building on the insights gained from Neural ODEs, which explore the depth of DNNs as a continuous variable, in this work, we generalize the traditional fully connected DNN through the concept of continuous width.
In the Generalized Deep Neural Network (GDNN), the traditional notion of neurons in each layer is replaced by a continuous state function. 
Using the finite rank parameterization of the weight integral kernel, we establish that GDNN can be obtained by employing the Integral Activation Transform (IAT) as activation layers within the traditional DNN framework. 
The IAT maps the input vector to a function space using some basis functions, followed by nonlinear activation in the function space, and then extracts information through the integration with another collection of basis functions. 
A specific variant, IAT-ReLU, featuring the ReLU nonlinearity, serves as a smooth generalization of the scalar ReLU activation. Notably, IAT-ReLU exhibits a continuous activation pattern when continuous basis functions are employed, making it smooth and enhancing the trainability of the DNN. Our numerical experiments demonstrate that IAT-ReLU outperforms regular ReLU in terms of trainability and better smoothness.
\end{abstract}

\tnotetext[fn1] {This manuscript has been authored by UT-Battelle, LLC, under contract DE-AC05-00OR22725 with the US Department of Energy (DOE). The US government retains and the publisher, by accepting the article for publication, acknowledges that the US government retains a nonexclusive, paid-up, irrevocable, worldwide license to publish or reproduce the published form of this manuscript, or allow others to do so, for US government purposes. DOE will provide public access to these results of federally sponsored research in accordance with the DOE Public Access Plan.}
\end{frontmatter}

\section{Introduction}
Deep learning, particularly deep neural networks (DNNs), has not only achieved remarkable success in traditional computer science fields such as computer vision \cite{he2016deep} and natural language processing \cite{devlin2018bert} but has also gained rapid popularity in other scientific communities, such as numerical partial differential equations (PDEs) \cite{raissi2019physics} and biology \cite{jumper2021highly}. Their cross-disciplinary popularity stems from their remarkable ability as highly expressive black-box function approximators. With high enough expressive power, as long as we can define a desired input-output map by a loss function, they can approximate such functions through brute force optimization, making them useful in many applications.

Despite the work of Hornik showing the universal approximation ability of DNNs with one hidden layer \cite{hornik1991approximation}, in practice, selecting an appropriate architecture with good hyperparameters, such as width and depth, is crucial. The architecture choice plays a significant role in achieving good practical expressive power, which refers to the model's trainability through gradient-based optimization \cite{schoenholz2016deep,hanin2018neural}. In other words, even with a large number of parameters, a poorly chosen architecture can result in limited practical expressive power, especially for deep architectures. By considering the model depth as a continuous variable, Neural ODEs, as introduced in \cite{chen2018neural}, offer improved expressive power without increasing the number of parameters. In this continuous depth setting, the forward propagation is formulated as an integral with respect to the depth variable.

Inspired by the idea of continuous depth, in this work, we introduce a General Deep Neural Network (GDNN) that explores the concept of continuous width. In GDNN, the state vector is regarded as a (continuous) function defined over the interval [-1,1]. Consequently, the weight matrices and bias vectors are also generalized to weight integral kernels and bias functions. The forward propagation of GDNN becomes recursively integrating the activated state functions with the weight kernel. A normal DNN can be viewed as a discretization of GDNN, where the standard weight matrices and bias vectors are obtained by evaluating the weight integral kernel and bias function on 2D and 1D mesh points, respectively. There are many existing methods to parameterize the weight kernel \cite{kovachki2021neural}, and in this paper, we mainly focus on the finite rank parameterization of the weight kernel. Under this parameterization, GDNN is equivalent to a traditional DNN with the Integral Activation Transform (IAT), which serves as a multivariate ($\bR^d \to \bR^d$)  nonlinear activation layer. In IAT, the input vector is first transformed into a state function by treating its elements as coefficients of some predetermined basis functions. Then, a pointwise activation is applied to the state function. Finally, the output vector is obtained by integrating the activated state function with another set of basis functions. We also show that by choosing rectangular functions as the basis functions, the IAT simplifies to the standard element-wise activation.

When the pointwise nonlinear activation is ReLU, we obtain a variant termed IAT-ReLU. Intriguingly, it is observed that both IAT-ReLU and element-wise ReLU exhibit a common characteristic where the activation matrix in the forward map is identical to the gradient. For example, applying ReLU element-wise to an input vector $\bz=[z_1,...,z_d]^T \in \bR^d$ can be expressed as $\phi(\bz) = \text{diag}([\textbf{1}_{z_1>0}(\bz),...,\textbf{1}_{z_d>0}(\bz)])\bz$, and the activation matrix $\text{diag}([\textbf{1}_{z_1>0}(\bz),...,\textbf{1}_{z_d>0}(\bz)])$ in the forward map is also the gradient. This connection can be further elucidated through Euler's theorem for homogeneous functions, as both ReLU and IAT-ReLU are classified as homogeneous functions.

At first, it might seem unsurprising that IAT-ReLU becomes equivalent to ReLU when employing rectangular basis functions. However, our findings indicate that this equivalence persists regardless of the basis functions chosen, effectively expanding the scope for developing ReLU-like activation functions beyond the limitations of rectangular basis functions. In IAT-ReLU, the activation matrix is continuously determined by the activation pattern $\mD(\bz)$, a compact subset of the interval $[-1,1]$. This activation pattern is defined as $\mD(\bz) = \{s\in[-1,1]: \bp^T(s) \bz \geq 0 \}$, where $\bz$ represents the input vector and $\bp(s)=[p_1(s),\ldots,p_d(s)]^T$ is a set of selected basis functions. With rectangular basis functions, the activation pattern comprises fixed sub-intervals, each aligning with the support of a rectangular basis function. The inclusion of these intervals depends on the sign of $\bz$, leading to a piecewise constant activation pattern, which also makes the gradient of ReLU discontinuous. Conversely, when using continuous basis functions, the activation pattern in IAT-ReLU is not confined to predefined sub-intervals. Instead, it can form infinitely many subsets within the domain $[-1, 1]$. Furthermore, we demonstrate that the activation pattern can vary continuously with the input $\bz$, presenting multiple advantages:

\begin{itemize}
    \item A continuous pattern leads to a continuous gradient, making the IAT-ReLU smooth and facilitating more stable training compared to ReLU, which is non-smooth and has a discontinuous gradient.
    \item The gradient can directly flow through the activation pattern, enabling the direct training of the activation pattern itself. In contrast, ReLU lacks gradient flow through the pattern as it is determined by the sign of $\bz$.
    \item With a continuous pattern, it becomes possible to decouple the activation pattern from the model parameters while still maintaining a continuous function. In the case of ReLU, decoupling the activation pattern results in a discontinuous model.
\end{itemize}
Moreover, the selection of basis functions with high total variation can increase the variability of activation patterns in domain $[-1,1]$. This is analogous to increasing the number of linear pieces in the ReLU network, where each linear piece is defined as the collection of inputs with the same activation pattern\cite{hanin2019deep}, and it is believed that by having more linear pieces the ReLU network will have more expressive power\cite{arora2016understanding,daubechies2022nonlinear}.

We summarize the main contributions of this work as follows:
\begin{itemize}
    \item We introduce the concept of GDNN and establish an equivalence between IAT and GDNN when using a finite rank parameterization for the weight kernel.
    \item We derive explicit expressions for the forward and backward computations of IAT-ReLU, as well as numerical methods to approximate them.
    \item We propose a generalize the activation pattern $\mD(\bz)$ and analyze the continuity and variation of $\mD(\bz)$ under different choices of basis functions. 
    \item We provide numerical examples to demonstrate that IAT-ReLU has  higher practical expressive power compared to ReLU, and show the impact of basis function selection and discretization on IAT-ReLU.
\end{itemize}

The rest of this paper is organized as follows: Section \ref{sec:1.1} offers an overview of related literature. Section \ref{sec:gdnn} details the construction of the GDNN as an extension of the traditional DNN. In Section \ref{sec:IAT}, the IAT is introduced and its association with GDNN is explored. Section \ref{sec:IAT-ReLU} analyzes and compares IAT-ReLU with the scalar ReLU, focusing on aspects such as smoothness, activation patterns, and the decoupling of the activation patterns. Section \ref{sec:exp} presents a range of numerical experiments to assess the influence of basis function choices, and discretization strategy in terms of expressive power.  The paper concludes in Section \ref{sec:5} with final remarks and perspectives.

\subsection{Related works:}\label{sec:1.1}

\textbf{Infinitely Wide Neural Network: } At initialization, the behavior of a neural network (NN) in the limit of infinite width is equivalent to that of a Gaussian process (GP). This correspondence was initially noted in \cite{neal1996bayesian} for NNs with a single hidden layer and was later extended to multilayer NNs in \cite{lee2017deep,matthews2018gaussian}. Additionally, \cite{jacot2018neural} demonstrated that the training dynamics of infinite-width NNs can also be described by a kernel called the Neural Tangent Kernel (NTK), which was later extended to convolutional neural networks by \cite{arora2019exact}. After discretization, the proposed GDNN can be transformed into a DNN of arbitrary width. However, it is important to note that the notion of width in the discretized GDNN differs significantly from the concept of infinitely wide neural networks in kernel regimes. In the discretized GDNN, the width is derived from the evaluation of the weight integral kernel function on a 2D mesh grid, which leads to non-i.i.d. samples. This is in contrast to the GP regime, which relies on the Central Limit Theorem and assumes i.i.d. parameters. Similarly, the NTK regime is based on overparameterization, whereas the integral kernel in the GDNN is governed by only moderate-sized parameters. Therefore, the discretized GDNN does not align with either the GP regime or the NTK regime.

\textbf{Integral transform in DNN: } The concept of using integral transforms to represent hidden layers in DNNs has been explored in the context of Neural Operator \cite{kovachki2021neural, li2020fourier}, but there are notable distinctions between their approach and ours. Firstly, their focus is primarily on operator learning, specifically mapping PDE parameters to PDE solutions, whereas our work centers around utilizing integral transforms to analyze the expressive power of DNNs. Secondly, they primarily study different ways of characterizing the integral transform and find that Fourier Neural Operator (FNO) \cite{li2020fourier} works best. In contrast, we focus on the finite rank parameterization of the integral kernel and derive the equivalent IAT as an activation layer, which is now only characterized by the choice of basis functions. Lastly, our use of a 1D hidden state function provides greater analytical tractability. For instance, with the ReLU activation function, we are able to compute the integral transforms analytically, enhancing the interpretability of the model.

\textbf{Expressive power of ReLU network: } The piecewise linear behavior of ReLU networks has been extensively studied, with research investigating convergence\cite{allen2019convergence, li2017convergence, du2018gradient, zhang2019learning, zou2018stochastic}, generalization ability\cite{allen2019learning, li2018learning, arora2019fine, zhang2019learning}, and expressive power\cite{allen2019learning, arora2019fine, lu2017expressive, daubechies2022nonlinear}. The linear regions in ReLU networks correspond to inputs with the same activation pattern, and estimating the maximum number of linear pieces has been a focus of a series of works\cite{pascanu2013number,montufar2014number,arora2016understanding,serra2018bounding,serra2020empirical}. Empirical evidence\cite{hanin2019deep, hanin2019complexity}, however, suggests that the practical number of linear regions is often lower than the theoretical maximum and exhibits minimal changes during training. In our proposed IAT-ReLU network, which becomes a piecewise linear function after discretization, we have control over the number of linear pieces through the choice of discretization. Analytically evaluating IAT-ReLU leads to infinitely many linear regions. Through numerical experiments, we have observed two effects of the number of linear pieces controlled by the mesh size. Firstly, it enhances solution smoothness, thereby improving performance in tasks that require smoothness, such as function fitting. Secondly, a larger number of linear pieces promotes gradient continuity and increases training stability.

\section{Generalized Deep Neural Network (GDNN)}\label{sec:gdnn}
In this section, we present the GDNN as a generalization of the traditional DNN, conceptualized by treating the network width as a continuous variable. Subsection \ref{sec:gdnn-construct} details the construction of the GDNN, and in subsection \ref{sec:gdnn-param}, we provide a parameterization framework for the GDNN.

\subsection{The Construction of GDNN}\label{sec:gdnn-construct}
We begin by constructing the GDNN as a continuous extension of the traditional DNN, from the perspective of width. This concept is illustrated through a basic 3-layer NN, comprising an input layer, an output layer, and one fully connected (FC) layer. These components encapsulate all necessary elements, and deeper structures can be easily achieved by stacking additional FC layers without further complications.

Consider $\bx \in \mathbb{R}^{d_0}$ as the model input and $\by \in \mathbb{R}^{d_3}$ as the NN output. The traditional NN structure is outlined as:
\begin{subequations}\label{eq:DNN}
\begin{align}
    \text{Input layer: }\;\;\;\;\; &\bz^{1} =  W^{0}\bx + \bb^{0}\\ 
    \text{FC layer: } \;\;\;\;\;\;\;\;\; &\bu^{1}=\sigma(\bz^{1}), \\
    &\bz^{2} = W^{1} \bu^{1} + \bb^{1},  \\
    \text{Output layer:}\;\;\;\;&\bu^{2} = \sigma(\bz^{2}),\\
    &\by = W^{2} \bu^{2} + \bb^{2},
\end{align}
\end{subequations}
where $W^n \in \mathbb{R}^{d_{n}\times d_{n-1}}$ for  $i=0,1,2$ are weight matrices, $\bb^n \in \mathbb{R}^{d_{n+1}}$, for $i=0,1,2$ are bias vectors and $\sigma(\cdot)$ is the nonlinear activation function.

Inspired by Neural ODEs \cite{chen2018neural}, which conceptualize depth as a continuous variable, our approach uniquely considers network width as a continuous variable. In line with this, the vectors of hidden neurons in the traditional DNN are replaced by state functions, defined within the interval $[-1,1]$. To accommodate for this change, the GDNN employs integral transforms instead of the affine transformations found in the traditional DNNs. In the input layer of GDNN, the input vector $\bx$ is used to expand the input weight functions $\boldsymbol{w}^{0}(s^1)$, creating a continuous state function $z^1(s^1)$ over the interval $[-1,1]$. The FC layer then applies a pointwise activation $\sigma(\cdot)$ to $z^1(s^1)$, resulting in the activated state function $u^1(s^1)$. This function is then integrated with the weight kernel $w^1(s^2, s^1)$ over $s^1$, generating the next state function $z^2(s^2)$. In the output layer, the activated state function $u^2(s^2)$ integrates with the output weight functions $\boldsymbol{w}^{2}(s^2)$, to yield the final output vector $\by$. Additionally, bias functions $b^0(s^1)$ and $b^1(s^2)$ are incorporated into the states $z^1(s^1)$ and $z^2(s^2)$, respectively. The complete GDNN structure is thus expressed as follows:

\begin{subequations}\label{eq:GDNN original}
\begin{align}
    \text{Input layer: }\;\;\;\;\; &z^{1}(s^{1}) =  \bx^T \boldsymbol{w}^{0}(s^{1}) + b^{0}(s^{1})\label{eq:4a}\\ 
    \text{FC layer: }\;\;\;\;\;\;\;\;\; & u^{1} (s^{1}) = \sigma(z^{1}(s^{1})), \\
    &z^{2}(s^{2}) = \int_{-1}^1  w^{1}(s^{2},s^{1}) u^{1}(s^{1})d s^{1} + b^{1}(s^{2}) , \label{eq:4c} \\
    \text{Output layer:}\;\;\;\;&u^{2} (s^{2}) =\sigma (z^{2}(s^{2})),\\
    &\by = \int_{-1}^1  \bw^{2} (s^{2}) u^{2}(s^{2})ds^{2} + \bb^{2} ,\label{eq:4e}
\end{align}
\end{subequations}
where $z^1,z^2:[-1,1] \to \mathbb{R}$ are the state functions, $u^1,u^2:[-1,1]\to \mathbb{R}$ are the corresponding activated state function, $\boldsymbol{w}^{0}: [-1,1]\to \mathbb{R}^{d_0}$ is the input weight function, $w^{1}: [-1,1]\times[-1,1] \to \mathbb{R}$ are weight kernel, $\boldsymbol{w}^{2}: [-1,1]\to \mathbb{R}^{d_3}$ is the output weight function,  $b^0, b^1: [-1,1] \to \mathbb{R}$ are bias functions and $\bb^{2}$ is the same bias vector in Eqs.(\ref{eq:DNN}).

\begin{figure}[h!]
  \centering
  {\includegraphics[width=0.99\textwidth]{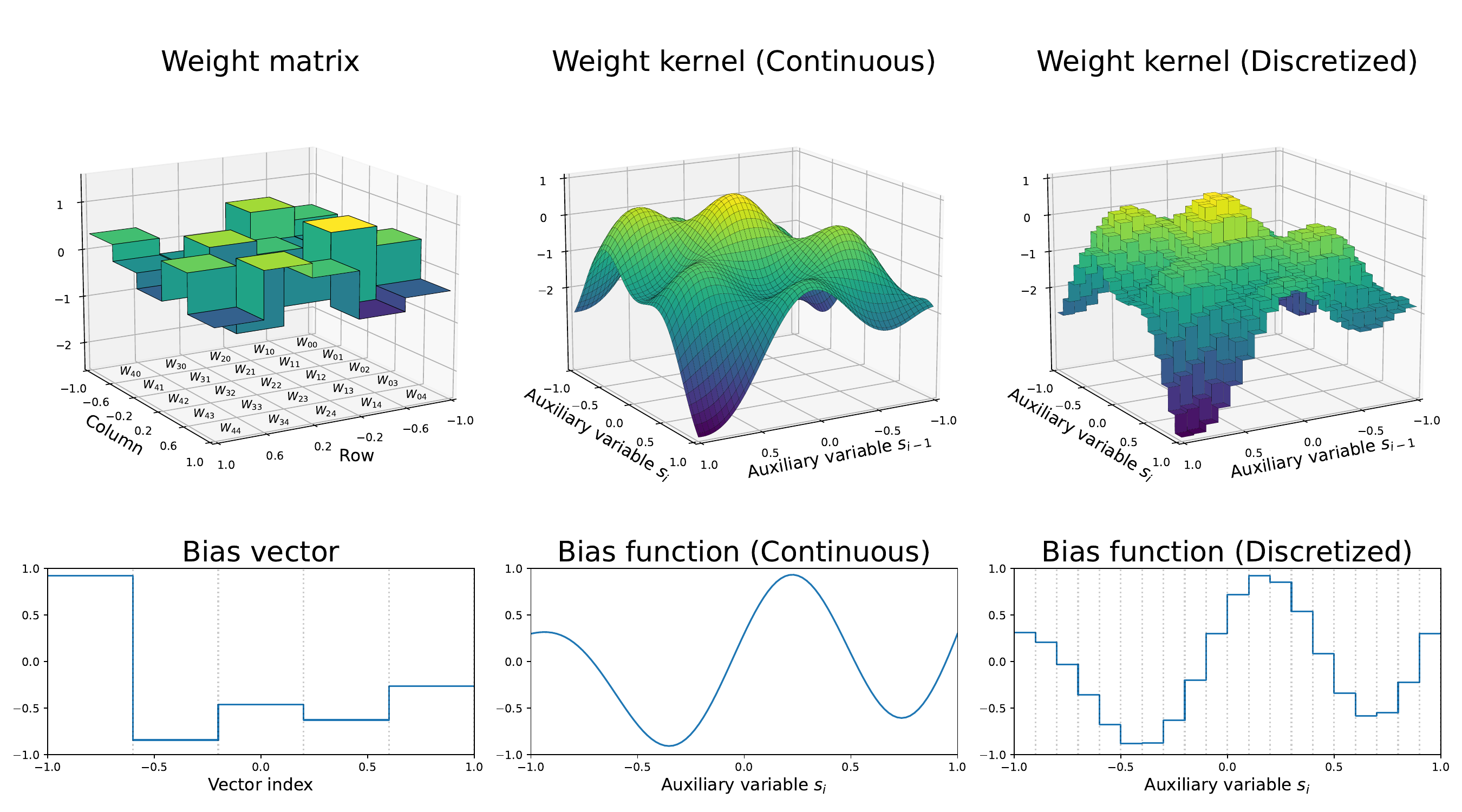}}
  \vspace{-0.1cm}
  \caption{The visualization for the weight kernels and bias functions. 
  Left column: The PC weight kernel and bias function, with the height of each segment controlled by the corresponding weight matrix and bias vector values. Middle column: The continuous weight kernel and bias function, as a generalization of the PC weight kernel and bias function. Right column: The discretized version of the continuous weight kernel and bias function, represented in the form of PC functions.}
  \label{fig:weight-vis}
\end{figure}

\textbf{DNN as a Special Case of  GDNN:} By selecting the 2D weight kernel $w^1(\cdot, \cdot)$, the 1D bias functions $b^0(\cdot)$ and $b^1(\cdot)$, and the 1D input/output weight functions $\boldsymbol{w}^{2}(\cdot)$ and $\boldsymbol{w}^{0}(\cdot)$ in Eqs. \eqref{eq:GDNN original} as piecewise constant (PC) functions, as illustrated in the left column 
of Figure\ref{fig:weight-vis}, we can exactly compute the integral transform in the GDNN. In such scenarios, this integral transform is equivalent to an affine transformation, simplifying the GDNN to a traditional DNN. In other words, the traditional DNN can be perceived as a particular instance of the GDNN, realized by selecting the generalized weight kernels, bias functions, and weight functions as PC functions.

\textbf{GDNN as an Infinitely Wide DNN:}
In cases where the weight kernel and bias functions are not piecewise constant, exact computation of the GDNN becomes generally infeasible, necessitating an appropriate discretization approach. In our framework, discretized weight matrices and bias vectors are derived by sampling the weight integral kernels and bias functions at the predetermined mesh points, resulting in matrices in $\mathbb{R}^{M \times M}$ and vectors in $\mathbb{R}^{M}$, with $M$ being the mesh size. After the discretization, integral transforms in GDNN become affine transformations, rendering the discretized GDNN analogous to a traditional DNN equipped with such affine transformations. Notably, the size of these affine transformations depends on the mesh size $M$, contrasting with traditional DNNs where affine transformations correspond to the fixed number of neurons in each layer. As $M$ tends towards infinity, the discretized GDNN progressively mirrors a finitely wide DNN, ultimately converging to the GDNN in its analytical form. This convergence can also be understood as the approximation of a continuous function using PC functions, as illustrated in the right column of Figure \ref{fig:weight-vis}.

\subsection{Parameterization of GDNN}\label{sec:gdnn-param}
In the GDNN framework, much like in traditional DNNs, the trainable parameters include the weight functions/kernels, $\bw^0(\cdot)$, $w^1(\cdot,\cdot)$, $\bw^2(\cdot)$, and bias functions/vectors $b^0(\cdot)$, $b^1(\cdot)$, $\bb^2$. However, the direct training of these infinite-dimensional functions is impractical, and it is necessary to adopt a suitable parameterization for effective training. 
In our approach, we use 1D basis functions to construct and parameterize all the continuous 1D weight and bias functions.  Additionally, the 2D weight integral kernels are formed by directly multiplying these 1D basis functions, a technique known as finite rank parameterization for the integral kernel.
The detailed parameterization scheme is as follows:

\begin{subequations}\label{eq:weight fun param}
\begin{align}
    \text{Input layer: }\;\;\;\;\; &\bw^{0}(s^{1}) =  (W^{0})^T \bp^{1}(s^{1})\label{eq:5a}\\
    &b^{0}(s^{1}) =(\bb^{0})^T \bp^{1}(s^{1}),\label{eq:5b}\\
    \text{FC layer: }\;\;\;\;\;\;\;\;\; & w^{1}(s^{2},s^{1}) = (\bp^{2}(s^{2}))^T W^1 \bq^1(s^1),\label{eq:5c}\\
    &b^{1}(s^{2}) = (\bb^{1})^T \bp^{2}(s^{2}), \label{eq:5d} \\
    \text{Output layer:}\;\;\;\;&\bw^{2} (s^{2}) = W^2 \bq^2(s^2),\label{eq:5e}
\end{align}
\end{subequations}
where $\{W^0, W^1, W^2,\bb^1,\bb^2\}$ are parameters from the same space as those in the DNN described in Eq.(\ref{eq:DNN}), and $\bp^{n},\bq^{n}: \mathbb{R} \to \mathbb{R}^{d_n}$ for $n=1,2$ are the selected collections of 1D input and output basis functions, respectively. In subsequent sections, we demonstrate that this parameterized GDNN is equivalent to a DNN equipped with a special Integral Activation Transform as its activation layers.

\section{Integral Activation Transform (IAT)}\label{sec:IAT}
In this section, we first give the definition of the IAT in Subsection \ref{sec:IAT-def}, then establish its connection with the GDNN in Subsection \ref{sec:IAT-GDNN}, and finally, present the numerical approximation of the IAT in Subsection \ref{sec:IAT-approx}.

\subsection{Integral Activation Transform (IAT)}\label{sec:IAT-def}
Let $\bp(\cdot) = [p_1(\cdot), ..., p_{d_1}(\cdot)]^T$ and $\bq(\cdot) = [q_1(\cdot), ..., q_{d_2}(\cdot)]^T$ denote two sets of basis functions, serving as the input and output basis functions for the IAT respectively, where $p_1,..., p_{d_1},q_1,...,q_{d_2}: [-1,1]\to \mathbb{R}$ are all 1D basis functions. We construct the IAT $\mI^{\sigma}_{\bp,\bq}: \mathbb{R}^{d_1} \to \mathbb{R}^{d_2}$ as a nonlinear activation layer in the following manner:

\begin{equation}\label{eq:IAT-integral}
    \bu = \mI^{\sigma}_{\bp,\bq} (\bz) = \int_{-1}^1 \bq(s) \sigma(\bz^T \bp(s)) ds 
\end{equation}
where $\sigma: \mathbb{R} \to \mathbb{R}$ represents the nonlinear activation function, with $\bz \in \mathbb{R}^{d_1}$ as the input and $\bu \in \mathbb{R}^{d_2}$ as the output.

\begin{figure}[h!]
  \centering
  {\includegraphics[width=0.8\textwidth]{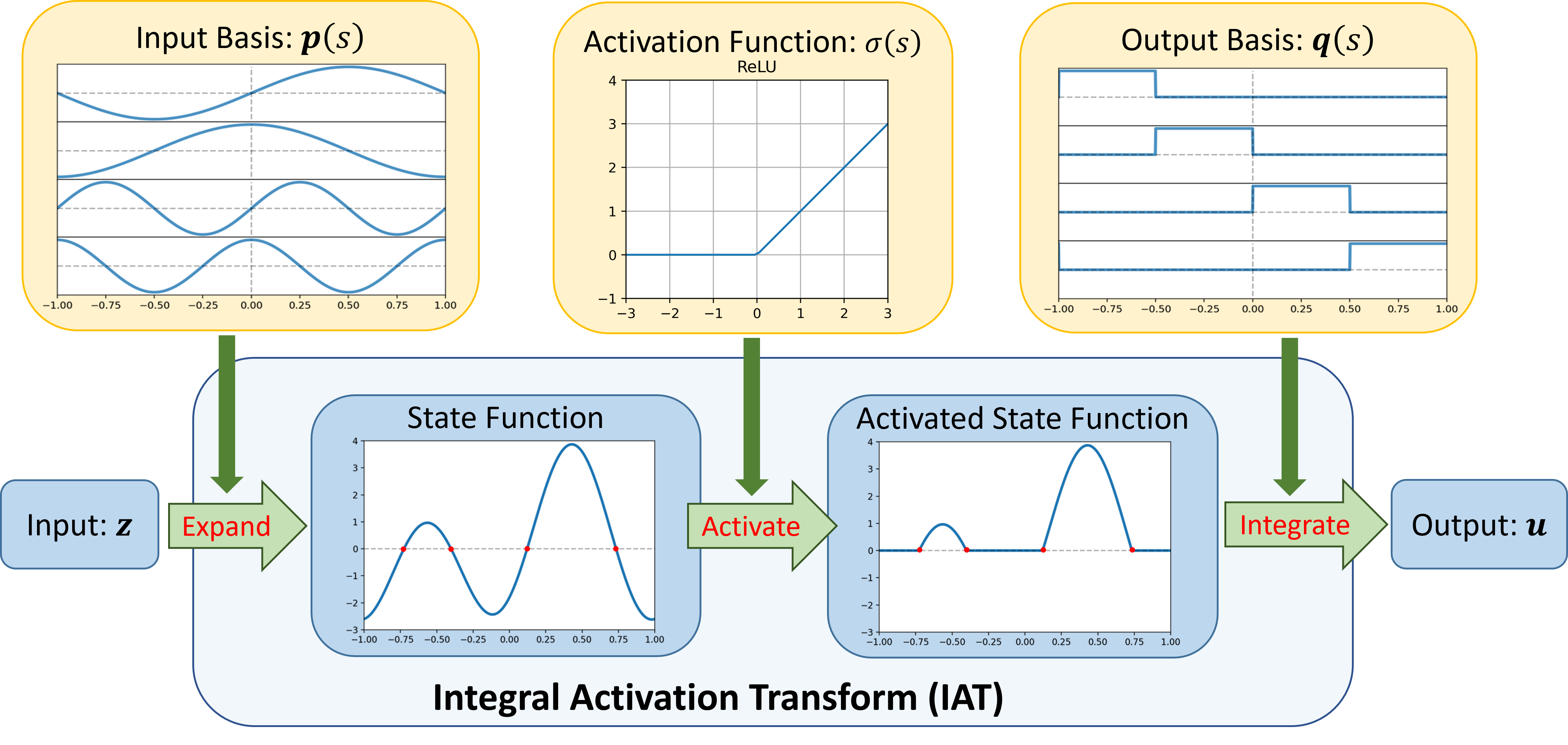}}
  \vspace{-0.1cm}
  \caption{The schematic for IAT with ReLU activation: First, the input vector $\bz$ is mapped to the state function $\bz^T \bp(s)$ by treating $\bz$ as coefficients of the input basis $\bp(s)$. Then, a pointwise activation function $\sigma$ is applied to the state function to introduce nonlinearity. Finally, the activated state function $\sigma(\bz^T \bp(s))$ is integrated with the output basis $\bq(s)$ to produce the output vector $\bu$. In the case of ReLU activation, the red dots represent the roots of the state function, which determine the segments of the state function that are set to zero.}
  \label{fig:IAT}
\end{figure}

Contrary to the typical activation functions in DNNs, which are applied to each component of the input $\bz$ to directly generate the components of the output $\bu$, the IAT adopts an alternative approach, operating activation in a function space defined over $[-1,1]$. In the IAT, the input vector $\bz$ is initially mapped to a state function $\bz^T \bp(s)$ defined on $[-1,1]$, with $\bz$ serving as the coefficients for the input basis $\bp(s)$. Then, the pointwise activation function $\sigma$ is applied to the state function to introduce the nonlinearity. Finally, this activated state function $\sigma(\bz^T \bp(s))$ is integrated with the output basis $\bq(s)$ to produce the output vector $\bu$. The conceptual diagram of IAT is depicted in Figure \ref{fig:IAT}.

\vspace{0.2cm}
\vspace{0.2cm}

In IAT, there is flexibility in selecting the input basis $\bp(\cdot)$, output basis $\bq(\cdot)$, and activation function $\sigma(\cdot)$. The input basis $\bp(\cdot)$ plays a crucial role in forming the state function $\bz^T \bp(s)$, upon which the nonlinear activation is applied. The choice of input basis significantly influences the interaction between the nonlinearity and the input vector $\bz$. Conversely, the output basis $\bq(\cdot)$ is in charge of extracting pertinent information from the activated state function, and different choices here can yield distinct IAT behaviors, a topic that will be explored in Section \ref{sec:exp-basis}.

A specific basis function worthy of mention is the rectangular function, depicted as the output basis $\bq(s)$ in Figure \ref{fig:IAT}. Rectangular functions represent the simplest form of PC functions. Selecting these rectangular functions for both $\bp(\cdot)$ and $\bq(\cdot)$ can lead to PC weight kernels and bias functions shown in Figure \ref{fig:weight-vis}. Consequently, this choice can simplify the IAT to a traditional element-wise activation function, described as:
\begin{equation}
\mI^{\sigma}_{\bp,\bq}(\bz) = \sigma(\bz),
\end{equation}
where $\sigma(\cdot)$ is individually applied to each component of $\bz$.

\subsection{Connection to GDNN}\label{sec:IAT-GDNN}

This subsection demonstrates the equivalence of the GDNN as described in Eqs. (\ref{eq:GDNN original}) with its parameterization in Eqs. (\ref{eq:weight fun param}) to a traditional DNN equipped with the IAT, defined in Eq. \eqref{eq:IAT-integral}. To proceed, we also assume that all considered basis functions $\bp^n(\cdot), \bq^n(\cdot), n=1,2$ are linear independent within each collection.

Firstly, in the input layer,  the state function $z^1(s^1)$ lies within the span of $\bp^1(s^1)$, and can be expressed as $$z^1(s^1) =  \left( \bp^1(s^1) \right)^T \bz^1,$$ where $\bz^1$ represents the coefficients for $\bp^1(s^1)$. By substituting Eqs. (\ref{eq:5a},\ref{eq:5b}) into Eq. (\ref{eq:4a}) we have
\begin{align*}
    z^1(s^1) &= \left((W^0)^T \bp^1(s^1) \right)^T \bx+  \left(\bp^1(s^1) \right)^T \bb^0\\
    &= \left(\bp^1(s^1) \right)^T  \left(W^0 \bx + \bb^0 \right).
\end{align*}
By linear Independence of $\bp^1(s^1)$, we obtain 
\begin{equation}\label{eq:GDNN-equiv1}
    \bz^1 = W^0 \bx + \bb^0.
\end{equation}

In the FC layer, $z^2(s^2)$ is within the span of $\bp^2(s^2)$ and can be represented as $$z^2(s^2) =  \left(\bp^2(s^2)\right)^T \bz^2.$$ By substituting Eqs. (\ref{eq:5c},\ref{eq:5d}) into Eq.(\ref{eq:4c}), we have 
\begin{align*}
    z^2(s^2) &= \int_{-1}^1 \left( \bp^2(s^2) \right)^T W^1 \bq^1(s^1) u^1(s^1) d s^1+ \left( \bp^2(s^2) \right)^T \bb^1\\
    &= \left( \bp^2(s^2) \right)^T  \left( W^1 \int_{-1}^1  \bq^1(s^1) u^1(s^1) d s^1   + \bb^1\right)\\
    &= \left( \bp^2(s^2) \right)^T \left( W^1 \int_{-1}^1  \bq^1(s^1) \sigma(( \bp^1(s^1) )^T \bz^1) d s^1   + \bb^1\right)\\
    &= \left( \bp^2(s^2) \right)^T \left( W^1 \mI_{\bp^1,\bq^1}^{\sigma}(\bz^1)  + \bb^1\right)\\
    &= \left( \bp^2(s^2) \right)^T \left( W^1 \bu^1  + \bb^1\right).
\end{align*}
Following the linear independence of $\bp^2(s^2)$, we obtain
\begin{equation}\label{eq:GDNN-equiv2}
    \bz^2 = W^1 \bu^1  + \bb^1,
\end{equation}
where 
\begin{equation}\label{eq:GDNN-equiv3}
    \bu^1 = \mI_{\bp^1,\bq^1}^{\sigma}(\bz^1).
\end{equation}

For the output layer, substituting Eq. (\ref{eq:5e}) into Eq.(\ref{eq:4e}) yields
\begin{align*}
    \by &=  \int_{-1}^{1} W^2 \bq^2(s^2) u^2(s^2) ds^2 + \bb^2 \\
    &=W^2 \int_{-1}^{1} \bq^2(s^2) \sigma(z^2(s^2)) ds^2 + \bb^2 \\
    &= W^2 \int_{-1}^{1}  \bq^2(s^2) \sigma((\bp^2(s^2))^T \bz^2) ds^2 + \bb^2 \\
    &= W^2 \int_{-1}^{1} \bq^2(s^2) \sigma((\bp^2(s^2))^T \bz^2) ds^2 + \bb^2 \\
    &= W^2 \mI_{\bp^2,\bq^2}^{\sigma}(\bz^2) + \bb^2 \\
    &= W^2 \bu^2 + \bb^2 .
\end{align*}
This gives the last piece of the derivation, and we have 
\begin{equation}\label{eq:GDNN-equiv4}
    \by = W^2 \bu^2  + \bb^2 ,
\end{equation}
where
\begin{equation}\label{eq:GDNN-equiv5}
    \bu^2 = \mI_{\bp^2,\bq^2}^{\sigma}(\bz^2).
\end{equation}

Combining the above derivations in Eqs. (\ref{eq:GDNN-equiv1},\ref{eq:GDNN-equiv2},\ref{eq:GDNN-equiv3},\ref{eq:GDNN-equiv4},\ref{eq:GDNN-equiv5}), the GDNN as described in Eqs.(\ref{eq:GDNN original}) with its parameterization in Eqs.(\ref{eq:weight fun param}) can be reformulated as follows:

\begin{subequations}\label{eq:GDNN activation}
\begin{align}
    \text{Input layer: }\;\;\;\;\; &\bz^{1} =  W^{0}\bx + \bb^{0} \label{eq:6a}\\ 
    \text{FC layer: }\;\;\;\;\;\;\;\;\; &\bu^{1}= \mI_{\bp^1, \bq^1}^{\sigma}(\bz^1), \label{eq:6b}\\
    &\bz^{2} = W^{1} \bu^{1} + \bb^{1},  \label{eq:6c}\\
    \text{Output layer:}\;\;\;\;&\bu^{2} = \mI_{\bp^2, \bq^2}^{\sigma}(\bz^2), \label{eq:6d}\\
    &\by = W^{2} \bu^{2} + \bb^{2},\label{eq:6e}
\end{align}
\end{subequations}
which is equivalent to the classical DNN in Eqs.(\ref{eq:DNN}) with the element-wise activation function $\sigma(\cdot)$ replaced by IAT $\mI_{\bp^n, \bq^n}^{\sigma}$ for $n=1,2$.

\subsection{The approximation of IAT}\label{sec:IAT-approx}

The IAT, $\mI^{\sigma}_{\bp,\bq}(\bz)$, as defined in Eq.(\ref{eq:IAT-integral}) comprises $d_2$ 1D integrals over the interval $[-1,1]$ and each integral is given by $\int_{-1}^1 q_i(s) \sigma(\bz^T \bp(s)) ds$, where $i=1,..,d_2$. Without making further assumptions on $\bq(\cdot)$, $\bp(\cdot)$, and $\sigma(\cdot)$, these integrals can be approximated using the midpoint rule with a uniform partition of $[-1,1]$. This discretized version of IAT, denoted as $\hat{\mI}^{\sigma}_{\bp,\bq}$, is then expressed as:
\begin{equation}\label{eq: IAT-numerical}
    \hat{\bu} = \hat{\mI}^{\sigma}_{\bp,\bq}(\bz) = \frac{2 Q \sigma(P^T \bz)}{M}
\end{equation}
where $M$ is the number of partition for the interval $[-1,1]$ and $P\in \bR^{d_1 \times M}$ and $Q\in \bR^{d_2 \times M}$ represent the evaluation of basis functions $\bp(\cdot)$ and $\bq(\cdot)$ at the $M$ chosen mesh points. The discretization of the IAT mirrors the discretization process of the weight kernel in GDNN, as discussed in Section \ref{sec:gdnn-construct}. Specifically, the discretized weight kernel $w^1(\cdot, \cdot)$, as parameterized in Eq. \eqref{eq:5c}, is expressed as an $M$ by $M$ matrix $\hat{W}^1 = (P^2)^T W^1 Q^1$, where  $P^2$ and $Q^1$ are matrices that correspond to the evaluations of the basis functions $\bp^2(\cdot)$ and $\bq^1(\cdot)$, respectively.

Furthermore, the gradient of the discretized IAT-ReLU, as defined in Eq.(\ref{eq: IAT-numerical}), can be computed as:
\begin{equation}\label{eq:IAT-gradient-num}
    \nabla_{\bz} \hat{\mI}^{\sigma}_{\bp,\bq}(\bz) = \frac{2 Q \;\text{diag}(\sigma'(P^T \bz)) \; P^T}{M}.
\end{equation}
According to \cite{folland1999real}, when $\sigma(\bz^T \bp(s))$ is differentiable for all $\bz$ and almost all $s \in [-1,1]$, the Leibniz integral rule can be applied to calculate the gradient of $\mI^{\sigma}_{\bp,\bq}(\cdot)$. By interchanging the integration and differentiation, the gradient is given by:
\begin{equation}\label{eq:IAT-gradient-ana}
    \nabla_{\bz} \mI^{\sigma}_{\bp,\bq}(\bz) =  \int_{-1}^1 \bq(s) (\bp(s))^T\sigma'(\bz^T \bp(s)) ds.
\end{equation}
This expression provides an analytical formula for the gradient, under the differentiability condition of $\sigma(\cdot)$. Notably, this analytical gradient aligns with the numerical gradient in Eq. (\ref{eq:IAT-gradient-num}). However, dealing with non-smooth activation functions like ReLU requires more careful consideration when interchanging integration and differentiation, a topic further explored in Section \ref{sec:IAT-ReLU}.

\section{IAT-ReLU}\label{sec:IAT-ReLU}
In this section, we focus on a variant of the IAT using the ReLU as the activation function, referred to as IAT-ReLU and denoted by $\mI_{\bp, \bq}^{\text{ReLU}}$. In Subsection \ref{sec:3.1}, we begin by establishing explicit formulas for the forward and backward computations in IAT-ReLU, demonstrating that IAT-ReLU generalizes the scalar ReLU. Subsequently, we investigate the activation patterns of IAT-ReLU in Subsection \ref{sec:3.2} and discuss its computational approach in Subsection \ref{sec:3.3}. Finally, we explore decoupling the activation pattern from model parameters in Subsection \ref{sec:3.4}.

\subsection{IAT-ReLU as a generalization of ReLU}\label{sec:3.1}
We begin by establishing that IAT-ReLU is a natural generalization of the scalar ReLU. 
For simplicity, we use $\phi(\cdot)$ to denote both scalar and element-wise ReLU, depending on the input type.
When ReLU is applied element-wise to a vector $\bz = [z_1, \ldots, z_d] \in \bR^d$, the operation can be expressed in the linear form as follows:
\begin{equation}\label{eq:ReLU-linear-form}
    \phi(\bz) = [\textbf{1}_{z_1\ge0}(\bz)z_1,...,\textbf{1}_{z_d\ge0}(\bz) z_d]^T = S^{\phi}(\bz) \bz
\end{equation}
where $S^{\phi}(\bz) = \text{diag}([\textbf{1}_{z_1\ge0}(\bz),...,\textbf{1}_{z_d\ge0}(\bz)]) \in \bR^{d \times d}$ is called the activation matrix. This activation matrix is piecewise constant, implying that $\nabla_{\bz} S^{\phi}(\bz) = 0$ almost everywhere. And it follows that the gradient of the element-wise ReLU is also equivalent to its activation matrix, expressed as:
\begin{equation}\label{eq:ReLU-gradient}
    \nabla_{\bz} \phi(\bz) =  S^{\phi}(\bz).
\end{equation}

We proceed to establish that IAT-ReLU exhibits similar properties to the scalar ReLU. For IAT-ReLU, we can rewrite its expression as follows:
\begin{align*}
    \mI_{\bp,\bq}^{\text{ReLU}}(\bz) &= \int_{-1}^{1} \bq(s) \sigma(\bz^T \bp(s)) ds\\
    &= \int_{-1}^{1} \bq(s) \textbf{1}_{\bz^T \bp(s) > 0}(s)(\bz^T \bp(s)) ds\\
    &= \int_{\{ s \in [-1,1]: \bz^T \bp(s) > 0\}} \bq(s) (\bp(s))^T  ds \bz\\
    &= S(\mD(\bz)) \bz. 
\end{align*}
This implies that the IAT-ReLU has the same linear form as follows:
\begin{equation}\label{eq:IAT-ReLU-linear}
     \mI_{\bp, \bq}^{\text{ReLU}}(\bz) = S(\mD(\bz)) \bz ,
\end{equation}
where $S(\mD(\bz)) \in \bR^{d \times d}$ is the activation matrix for IAT-ReLU. Its expression is given by   
\begin{equation}\label{eq:IAT-ReLU-activation-matrix}
    S(\mD(\bz)) = \int_{\{s\in \mD(\bz)\}} \bq(s) \bp^T(s) ds. 
\end{equation}
where $\mD(\bz)$ is called the activation pattern and is defined as:
\begin{equation}\label{eq:activation-pattern}
    \mD(\bz) = \{s\in [-1,1]: \bp^T(s) \bz > 0 \}.
\end{equation}
This formulation confirms that IAT-ReLU maintains the linear form analogous to the scalar ReLU.

We now derive the gradient of $\mI_{\bp,\bq}^{\text{ReLU}}(\bz)$. 
Let the state function be denoted as $f(s, \bz) = \bz^T \bp(s)$ and we can first rewrite $\mI_{\bp,\bq}^{\text{ReLU}}(\bz)$ as
\begin{align*}
    \mI_{\bp,\bq}^{\text{ReLU}}(\bz) = \int_{\mD(\bz)} \bq(s) f(s,\bz) ds.
\end{align*}
The dependence of $\mI_{\bp,\bq}^{\text{ReLU}}(\bz)$ on the input $\bz$ is through two paths: the state function $f(s,\bz)$ and the activation pattern $\mD(\bz)$.  The set $\mD(\bz)$, a subset of $[-1,1]$, consists of disjoint intervals characterized by boundary points $r_k, k=1,\ldots,K$, at which $f(\cdot, \bz)$ changes sign. Therefore, the gradient of $\mI_{\bp,\bq}^{\text{ReLU}}(\bz)$ will have the general expression as follows:
\begin{equation}\label{eq:IAT-relu-gradient-general}
    \nabla_{\bz} \mI_{\bp, \bq}^{\text{ReLU}}(\bz) = S(\mD(\bz)) + \sum_{k=1}^K f(r_k, \bz) \bq(r_k) (\nabla_{\bz} r_k)^T
\end{equation}
where $\nabla_{\bz} r_k$ is the gradient of boundary point $r_k$ with respect to the input $\bz$. In the above expression, the first term represents the gradient through $f(s,\bz)$ and the second sum stands for the gradient on $\mD(\bz)$ through each $r_k$.

While the dependence through $f(s,\bz)$ always exists, we only need to analyze the dependence through $\mD(\bz)$ via $r_k$, and we identify two scenarios for such dependency. In the first scenario, $f(\cdot,\bz)$ intersects the x-axis at $r_k$, making $r_k$ a root of $f(\cdot,\bz)$. Here, $r_k$ has a nonzero gradient with respect to $\bz$, i.e., $\nabla_{\bz} r_k \neq 0$. However, $f(r_k,\bz) = 0$ since $r_k$ is the root, resulting in $f(r_k, \bz) \bq(r_k) (\nabla_{\bz} r_k)^T = 0$. In the second scenario, $f(\cdot,\bz)$ changes sign at $r_k$ due to a jump discontinuity. From the definition of the $f(\cdot,\bz)$, the discontinuities in $f(\cdot,\bz)$ can only come from the discontinuities within basis $\bp(\cdot)$, and are independent of the value of $\bz$. Therefore, we have $\nabla_{\bz} r_k = 0$, which leads to  $f(r_k, \bz) \bq(r_k) (\nabla_{\bz} r_k)^T = 0$ despite $f(r_k, \bz) \neq 0$. Combining the both cases,  the gradient of $\mI_{\bp, \bq}^{\text{ReLU}}(\bz)$ is simplified to: 
\begin{equation}\label{eq:IAT-relu-gradient}
    \nabla_{\bz} \mI_{\bp, \bq}^{\text{ReLU}}(\bz) = S(\mD(\bz)).
\end{equation}
This implies that the gradient of $\mI_{\bp, \bq}^{\text{ReLU}}(\bz) = S(\mD(\bz)) \bz$ with respect to $\bz$ equals $S(\mD(\bz))$, despite the non-zero gradient of $S(\mD(\bz))$ on $\bz$.

By comparing Eq.\eqref{eq:ReLU-linear-form} with Eq.\eqref{eq:IAT-ReLU-linear}, and Eq.\eqref{eq:ReLU-gradient} with Eq.\eqref{eq:IAT-relu-gradient}, it becomes evident that both ReLU and IAT-ReLU exhibit a shared linear form, as characterized by their activation matrices, respectively. Additionally, their gradients are found to be equivalent to their activation matrices. The primary difference between them resides in the specific values taken by the activation matrix $S(\mD(\bz))$ as a function of $\bz$. The forthcoming subsections will undertake a detailed investigation of the behavior of the activation matrix $S(\mD(\bz))$.

In addition, the gradient formulation in Eq.\eqref{eq:IAT-relu-gradient}, derived for the non-smooth activation ReLU, aligns with the gradient in Eq.\eqref{eq:IAT-gradient-ana}, obtained under the assumption of a smooth activation function. This alignment suggests that the numerical approximation in Eq.\eqref{eq: IAT-numerical} remains consistent for IAT-ReLU as well. This consistency underscores the robustness of the IAT framework, capable of accommodating both smooth and non-smooth activation functions without compromising the integrity of its gradient approximation.

\subsection{The activation pattern of IAT-ReLU}\label{sec:3.2}
From Eq.\eqref{eq:IAT-ReLU-linear}, we observe that the nonlinearity within $\mI_{\bp, \bq}^{\text{ReLU}}(\bz)$ is determined by the variation of $S(\mD(\bz))$.
Specifically, when $S(\mD(\bz))$ remains locally constant, $\mI_{\bp, \bq}^{\text{ReLU}}(\bz)$ behaves as a linear map. 
Further inspection of  Eq. \eqref{eq:IAT-ReLU-activation-matrix} reveals that the activation matrix $S(\mD(\bz))$ is dependent on the input $\bz$ through the activation pattern $\mD(\bz)$.  
It is important to note that the function $S(\cdot)$ itself is not directly related to $\bz$.
Hence, the subsequent subsections focus on analyzing the variation of $\mD(\bz)$ as a function of $\bz$. 
We will investigate two distinct types of variations in $\mD(\bz)$: local variations and global variations.

\subsubsection{The local variation of $\mD(\bz)$}\label{sec:3.2.1}
The local variation of $\mD(\bz)$ concerns how $\mD(\bz)$ changes relate to local variations in $\bz$. 
Here, we specifically investigate two aspects of such local variation: the presence of local changes in $\mD(\bz)$ as $\bz$ varies locally and the continuity of such changes in $\bz$.

As discussed above, we know that $\mD(\bz)$ is a subset of $[-1,1]$, and there are two scenarios on how $\mD(\bz)$ is determined by $\bz$: 
\begin{itemize}
    \item In the first scenario, $\mD(\bz)$ is determined by the roots of $f(\cdot;\bz) = \bz^T \bp(\cdot)$, which requires $f(\cdot;\bz)$ to be continuous, and this is achieved by choosing continuous input basis functions $\bp(\cdot)$. 
    Generally speaking, variations in the coefficients $\bz$ induce perturbations in the roots of $f(\cdot; \bz)$, ultimately causing changes in $\mD(\bz)$. 
    Therefore, when $\mD(\bz)$ determined by the roots of $f(\cdot;\bz)$, there are local changes in $\mD(\bz)$ as $\bz$ changes.
    In addition, if the $r_k$ is a single root of $f(\cdot; \bz)$, the gradient is given by 
    \begin{equation}\label{eq:root-gradient}
        \frac{\partial r_k}{\partial \bz} = - \frac{\bp(r_k)}{\bz^T \bp'(r_k)}.
    \end{equation}
    If $\bp(\cdot)$ and $\bp'(\cdot)$ are both continuous and given that $\mD(\bz)$ is solely determined by single roots,  then $\mD(\bz)$ will be continuous in $\bz$. 
    In addition, if $\bp(\cdot)$ are analytic functions, the continuity of $\mD(\bz)$ in $\bz$ is ensured, regardless of the root types, owing to the fact that the roots of analytic function are continuous functions of their coefficients. 
    \item In the second scenario, $\mD(\bz)$ is determined by the discontinuity of $f(\cdot;\bz)$, which comes from the discontinuities in $\bp(\cdot)$. 
    The rectangular basis functions, illustrated as the output basis in Figure \ref{fig:IAT}, exemplify this situation.
    Since the discontinuities in $\bp(s)$ do not change, $\mD(\bz)$ is merely selecting the sub-intervals on which $f(\cdot;\bz)$ is positive, and the sub-intervals are defined by the fixed discontinuities in $\bp(s)$. 
    Consequently, $\mD(\bz)$ is confined to a finite set of possible choices.
    As $\bz$ changes, based on the active discontinuities that separate the positive and negative part of $f(\cdot;\bz)$, $\mD(\bz)$ either remains unchanged or undergoes an abrupt transition to another configuration.
    Thus, when $\mD(\bz)$ is shaped by the discontinuity of $f(\cdot;\bz)$, local changes in $\mD(\bz)$ from $\bz$ are absent except for jumps. 
    This stands in stark contrast to the first scenario, where $\mD(\bz)$ can undergo continuous changes as $\bz$ varies. Additionally, in this second scenario, it is obvious that $\mD(\bz)$ lacks continuity in $\bz$.
\end{itemize}

From the integral definition in Eq. \eqref{eq:IAT-ReLU-activation-matrix}, it can be seen that $S(\cdot)$ is absolute continuity with respect to $\mD(\bz)$. The aforementioned properties of $\mD(\bz)$ can be directly extended to characterize the activation matrix $S(\mD(\bz))$, which governs the gradient $\nabla_{\bz} \mI_{\bp, \bq}^{\text{ReLU}}(\bz)$, as expressed in Eq. \eqref{eq:IAT-relu-gradient}.

Regarding the mapping $\mI_{\bp, \bq}^{\text{ReLU}}(\bz)$ itself, the continuity of $\mD(\bz)$ dictates the smoothness of $\mI_{\bp, \bq}^{\text{ReLU}}(\bz)$, since the continuity of $\mD(\bz)$ decides the continuity of $S(\mD(\bz))$ in $\bz$. Therefore, a continuous $\mD(\bz)$ yields a smooth $\mI_{\bp, \bq}^{\text{ReLU}}(\bz)$, and vise versa.

In the context of training a NN with $\mI_{\bp, \bq}^{\text{ReLU}}(\bz)$, two major consequences arise. The first one is the plateau phenomenon observed during training, characterized by an initial rapid loss reduction, followed by an extended stagnation period before another rapid decrease. This behavior is commonly observed in NNs with ReLU activation in the literature. The plateau is a consequence of $S(\mD(\bz))$ being piecewise constant, which, in turn, results from the piecewise constant nature of $\mD(\bz)$.
If $\mD(\bz)$ remains static during training, the nonlinearity $S(\mD(\bz))$ remains unchanged as the model parameters are updated, effectively making the NN behave like a linear model.
Gradient descent can rapidly minimize the loss within the subspace defined by the fixed $S(\mD(\bz))$ \cite{ainsworth2021plateau}, leading to a quick initial loss decrease followed by a prolonged stagnation period until a more favorable $\mD(\bz)$ emerges, if at all. 
Continuous changes in $\mD(\bz)$ as the model parameters are updated prevent gradient descent from saturating within a fixed subspace, reducing the likelihood of the model getting stuck in a local minimum and thereby enhancing overall trainability.

The second consequence is training stability, a property derived from the continuity of the gradient $S(\mD(\bz))$, which is ultimately determined by the continuity of $\mD(\bz)$. When the gradient $S(\mD(\bz))$ is discontinuous, slight changes in the parameters could lead to significant jumps in the gradient, leading to unstable or divergent training dynamics. This instability is often manifested by a highly fluctuating loss curve. Common strategies to address this issue involve using a smaller learning rate, which, however, slows down training, or overparameterizing the model to control the expected changes in the activation pattern, aiming to stabilize the training process \cite{allen2019convergence}. In contrast, the use of analytic $\bp(\cdot)$ to achieve a continuous $\mD(\bz)$ ensures that the gradient remains continuous with respect to the model parameters, eliminating unexpected jumps. This continuity in the gradient facilitates a much smoother decay of the loss during the training process compared to scenarios with a discontinuous gradient.

\subsubsection{The global variation of $\mD(\bz)$}\label{sec:3.2.2}
The global variation of $\mD(\bz)$ assesses how much $\mD(\bz)$ varies across the input domain. 
In Eq. \eqref{eq:IAT-ReLU-linear},  the nonlinearity of $\mI_{\bp, \bq}^{\text{ReLU}}(\bz)$ is determined by changes in the activation matrix $S(\mD(\bz))$ via $\mD(\bz)$.
Specifically, if a ReLU network exhibits a constant activation pattern in a region, the model reduces to a linear map in that region, limiting its expressive power. 
Many researchers posit that the number of activation patterns (linear pieces) in a ReLU network contributes to its expressiveness \cite{montufar2014number}. 
To enhance a ReLU network's expressiveness, diverse activation patterns are preferred across the input space.
This implies that as the input varies, the activation pattern should change to introduce more linear pieces. 
With a greater number of distinct activation patterns corresponding to various input space regions, the network achieves enhanced expressive capacity to model intricate relationships and capture diverse data patterns.

Conversely, with the selection of a continuous input basis $\bp(\cdot)$, IAT-ReLU can manifest a continuous activation pattern, akin to finitely many linear pieces. Each linear piece represents a straight line passing through the origin, as the positive part of $f(\cdot; \bz)$ remains unchanged when the input $\bz$ is scaled to $a \bz$, where $a$ is a non-zero constant. However, an infinite number of linear pieces does not necessarily imply increased expressive power if the activation pattern only exhibits slight changes across the input domain. Therefore, choosing an appropriate $\bp(\cdot)$ is crucial to ensure that the activation pattern $\mD(\bz)$ explores a diverse range of subsets in the interval $[-1,1]$ as $\bz$ varies. As mentioned earlier, the activation pattern $\mD(\bz)$ is determined by the roots of $f(\cdot;\bz)$. In selecting basis functions, two strategies can be considered to enhance the variation of $\mD(\bz)$ by increasing the diversity of roots in $f(\cdot; \bz)$ as $\bz$ varies.

\begin{itemize}
    \item Use basis functions with zero integral: Using input basis functions $\bp(\cdot)$ each with zero integral ensures that the state function $f(\cdot, \bz)$ also possesses a zero integral. This guarantees that $f(\cdot, \bz)$ intersects the x-axis, leading to at least one root or two roots if $\bp(\cdot)$ are also cyclic. This condition prevents situations where the $f(\cdot, \bz)$ lies entirely above or below zero. Additionally, each $\bz$ will activate approximately half of $[-1, 1]$, resulting in a more balanced distribution of $\mD(\bz)$ as a subset of $[-1, 1]$.
    \item Use basis functions with large total variations: Another strategy is to select basis functions characterized by large total variations, with typical examples being global basis functions. These functions are sensitive to small changes in any component of the input $\bz$, inducing substantial variations in the state function $f(\cdot; \bz)$ as well as its roots. As a result, the activation pattern will demonstrate increased diversity across the input space, potentially granting the model greater nonlinearity to capture diverse patterns in the data and enhance its expressive power.
\end{itemize}

\subsection{The approximation of IAT-ReLU}\label{sec:3.3}

\begin{figure}[h!]
  \centering
  {\includegraphics[width=0.99\textwidth]{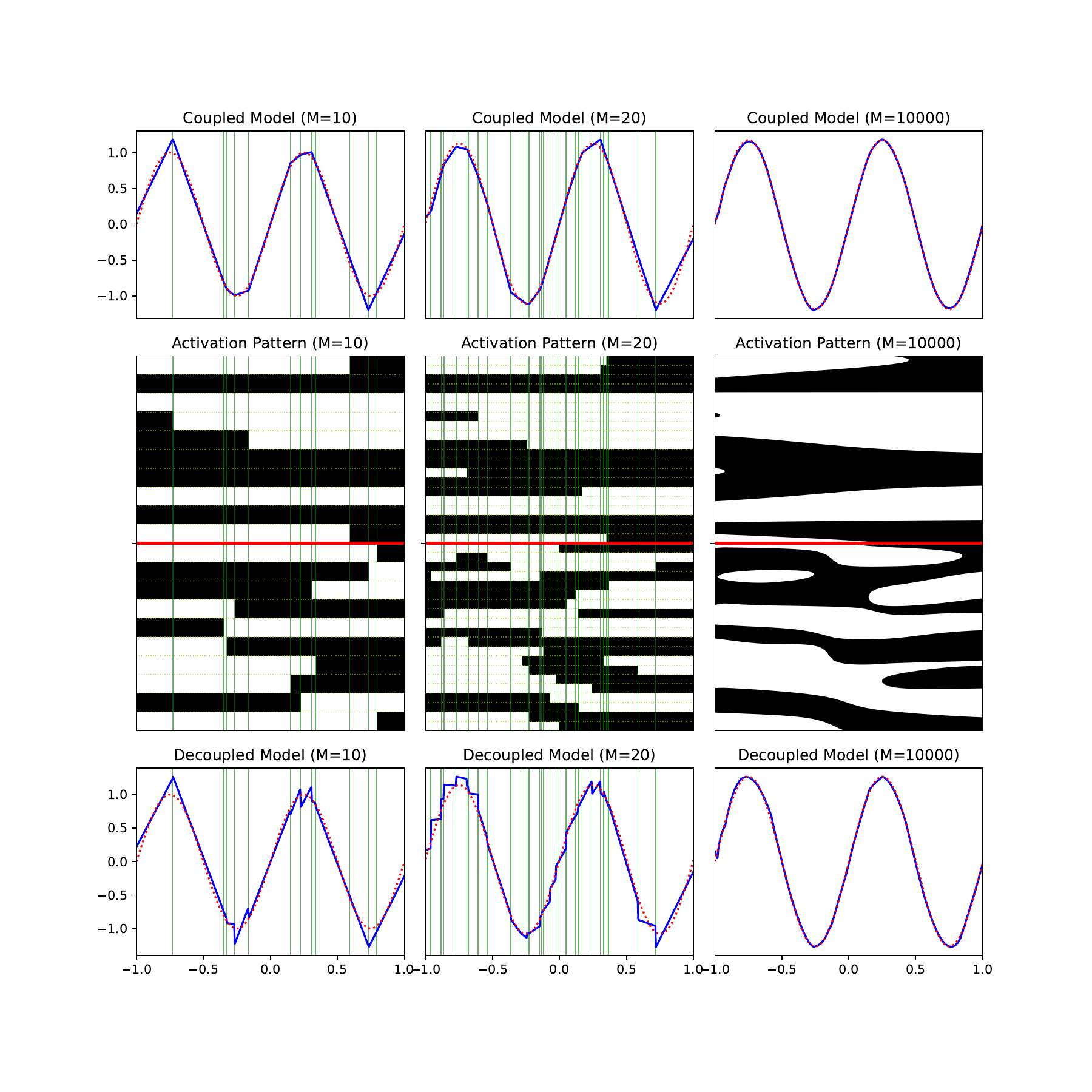}}
  \vspace{-0.8cm}
  \caption{Activation pattern for 3-layer IAT-ReLU network in Eqs.(\ref{eq:GDNN activation}) with 10 neurons in each layer for fitting $\sin(2\pi x)$ in $[-1,1]$. 
  Top row: The fitted solution for original IAT-ReLU the model. 
  Middle row: The activation patterns at the two hidden layers. 
  The x-axis corresponds to the input space, while the y-axis represents the two activation pattern spaces, with the first hidden layer positioned above the red line and the second hidden layer below the red line.
  Bottom row: The fitted solution for the decoupled model, where the activation pattern is taken from the trained model in the middle row. 
  From left to right, we show the results for M=10,20, and 10000, respectively. For M=10 and M=20, the locations where the activation pattern changes are marked by the green dashed lines and we consider M=10000 as an example of the continuous activation pattern. As we increase the mesh size M, each linear piece becomes smaller, and the fitted solution becomes smoother while still being a piecewise linear function. It becomes a truly smooth function when the activation pattern is continuous. For discrete patterns (M=10 and 20), the decoupled model exhibits discontinuities where the activation pattern changes. On the other hand, the continuous activation pattern still provides a continuous solution even after we decouple the activation pattern from the parameters that generate them. 
}
  \label{fig:actv-pt}
\end{figure}

The gradient of $\mI_{\bp, \bq}^{\text{ReLU}}(\bz)$ in Eq.(\ref{eq:IAT-relu-gradient}) is equivalent to the gradient of IAT for smooth activation action in Eq.(\ref{eq:IAT-gradient-ana}). Therefore, the discretization method used in Eq.(\ref{eq: IAT-numerical}) can also be applied to $\mI_{\bp, \bq}^{\text{ReLU}}(\bz)$, and we denote this discretized version as $\hat{\mI}_{\bp, \bq}^{\text{ReLU}}(\bz)$. Under this discretization, the activation pattern $\mD(\bz) \subset [-1,1]$ is represented by the discretized activation pattern $\hat{\mD}(\bz) \in \bR^M$, which captures the sign of $f(\cdot, \bz)$ on the M chosen evaluation points. As a result, $\hat{\mD}(\bz)$ remains piecewise constant, as very small perturbations in $\bz$ may not change the sign of $f(\cdot, \bz)$ on the evaluation points if the evaluation points are not dense enough. Therefore, $\hat{\mI}_{\bp, \bq}^{\text{ReLU}}(\bz)$ remains a piecewise linear function. However, by choosing a large number of evaluation points $M$, we can make each linear piece smaller and make $\hat{\mI}_{\bp, \bq}^{\text{ReLU}}(\bz)$ smoother while still being piecewise linear. As $M$ approaches infinity, we obtain infinitely many linear pieces, and $\hat{\mI}_{\bp, \bq}^{\text{ReLU}}(\bz)$ converges to $\mI_{\bp, \bq}^{\text{ReLU}}(\bz)$ which is a smooth function. An example of the above discussion is demonstrated in Figure \ref{fig:actv-pt}.

Alternatively, to circumvent the reliance on numerical integration techniques, we can approximate the IAT-ReLU computation nearly exactly by analytically solving the integral in Eq.(\ref{eq:IAT-integral}). This approach involves accurately finding the roots of the state function, which are illustrated as the red dots in Figure \ref{fig:IAT}. This can be accomplished either through root-finding algorithms like bisection or Newton's method, or by employing specific classes of basis functions. For instance, when using piecewise linear basis functions, the state function becomes piecewise linear, allowing for the roots to be analytically determined for each linear segment. Once these roots are identified, the integral can be computed analytically by applying the fundamental theorem of calculus. It is worth noting that the root-finding procedure can be performed off the computation graph during forward propagation since gradients from $\bz$ to roots are multiplied by zero in the gradient of $\mI_{\bp, \bq}^{\text{ReLU}}(\bz)$. By achieving high-accuracy root finding, we can obtain a continuous activation pattern $\mD(\bz)$ as a subset of $[-1,1]$, resulting in a smooth $\mI_{\bp, \bq}^{\text{ReLU}}(\bz)$. However, this approach requires more computational resources for the root-finding procedure compared to the computational cost of the two extra matrix multiplications involved in Eq.(\ref{eq: IAT-numerical}).

\begin{rem}[Higher order derivative]
For applications that require higher-order derivatives, such as Physics-Informed Neural Networks (PINN), it is necessary to compute $\mI_{\bp, \bq}^{\text{ReLU}}(\bz)$ analytically instead of using the discretized version $\hat{\mI}_{\bp, \bq}^{\text{ReLU}}(\bz)$, as the latter has zero second-order derivatives everywhere, similar to the ReLU activation. To capture higher-order derivatives, it is important to preserve the gradient between $\bz$ and the roots $r_k$ when computing $\mI_{\bp, \bq}^{\text{ReLU}}(\bz)$ analytically. This can be achieved by numerically inserting the value of the gradient from Eq.\eqref{eq:root-gradient} into the roots $r_k$ obtained from the root-finding algorithms.  Alternatively, if the basis functions $\bp(\cdot)$ are piecewise linear, it is possible to derive an explicit expression for the root in terms of $\bz$. However, this specific aspect is left as future work. In the scope of this paper, we focus on tasks that do not require higher-order derivatives, such as classification and regression.
\end{rem}

\begin{rem}[Scaling of the basis functions]
In addition to choosing the shape of the basis functions $\bp(\cdot)$ and $\bq(\cdot)$, it is also important to select a proper scaling constant for them. It is worth noting that $\mI_{\bp, \bq}^{\text{ReLU}}(\bz)$ exhibits homogeneity with respect to the scaling of $\bp(\cdot)$, $\bq(\cdot)$, and $\bz$. Let $a_1$, $a_2$ and $a_3$ be three scalars, the homogeneity of  $\mI_{\bp, \bq}^{\text{ReLU}}(\bz)$ means
\begin{equation}
    \mI_{a_1\bp, a_2\bq}^{\text{ReLU}}(a_3\bz) = a_1a_2a_3\mI_{\bp, \bq}^{\text{ReLU}}(\bz).
\end{equation}
This equation shows that scaling $\bp(\cdot)$, $\bq(\cdot)$, and $\bz$ does not change the overall nonlinearity of the function. One simple trick that can be used to avoid the scaling issue is to apply batch normalization after the IAT-ReLU layer. Batch normalization can help standardize the IAT-ReLU and make them more robust to scaling variations.
\end{rem}

\subsection{The decoupled IAT-ReLU network}\label{sec:3.4}
When considering the IAT-ReLU as an activation layer in a DNN, an interesting aspect is to decouple the activation matrices from the hidden states that generate them. For the purpose of demonstration, let's consider the same 3-layer DNN described in Eqs.(\ref{eq:DNN}). Using the linear form from Eq.(\ref{eq:IAT-ReLU-linear}), we can write the DNN in a similar deep linear form as follows:
\begin{equation}\label{eq:DNN-linear}
    \by = W^2(S(D^2) (W^1(S(D^1) (W^0 \bx + b^0)  + b^1)) + b^2
\end{equation}
where $D^1=\mD(\bz^1)$ and $D^2=\mD(\bz^2)$ is the activation patterns of hidden states $\bz^1$ and $\bz^2$, respectively. We can also write Eq.(\ref{eq:DNN-linear}) in a more general form as follows:
\begin{equation}\label{eq:DNN-coupled}
    \by = F(\bx, \textbf{W}, \boldsymbol{\mD}(\bx, \textbf{W})) 
\end{equation}
where $\textbf{W} = [W^0,b^0,W^1,b^1,W^2,b^2]$ represents all the parameters of the DNN, and $\boldsymbol{\mD}(\bx, \textbf{W}) = [\mD^1(\bx, \textbf{W}), \mD^2(\bx, \textbf{W})]$ represents all the activation patterns as a function of the model input $\bx$ and the parameter $\textbf{W}$.

Indeed, we can observe that the nonlinearity in the model is solely determined by the activation pattern $\boldsymbol{\mD}(\bx, \textbf{W})$. When $\boldsymbol{\mD}(\bx, \textbf{W})$ remains constant with respect to $\bx$ and \textbf{W}, the model output $\by$ exhibits a linear relationship with the input $\bx$ and each parameter pair $(W_n, b^n)$, where $n = 0, 1, 2$. Thus, the $\textbf{W}$ in the second argument of $F(\cdot, \cdot, \cdot)$ controls the linearity of the mapping from $\bx$ to $\by$, while the $\textbf{W}$ in $\boldsymbol{\mD}(\bx, \textbf{W})$ controls the nonlinearity from $\bx$ to $\by$.

For instance, in the case of a ReLU DNN, $\boldsymbol{\mD}(\bx, \textbf{W})$ characterizes each linear piece, and the \textbf{W} in the second position of $F(\cdot, \cdot, \cdot)$ determines the linear mapping within each piece. We also believe that using the same \textbf{W} to control both the linear and nonlinear aspects of the model simultaneously makes the optimization for \textbf{W} challenging. This is in contrast to the traditional approach where the mesh structure, which can be viewed as the linear pieces represented by $\boldsymbol{\mD}(\bx, \textbf{W})$, is first determined, and subsequent computations are performed within each linear piece, which is governed by the \textbf{W} in the second position of $F(\cdot, \cdot, \cdot)$.

Therefore, a natural idea is to decouple the linearity and nonlinearity of the model into two collections of parameters. In other words, we can rewrite Eq.(\ref{eq:DNN-coupled}) as follows:
\begin{equation}
    \by = F(\bx, \textbf{W}_1, \boldsymbol{\mD}(\bx, \textbf{W}_2)) 
\end{equation}
where $\textbf{W}_1$ is the parameters controlling the linearity and $\textbf{W}_2$ represents the parameters associated with the nonlinearity through the activation pattern.

This idea has been explored in the work of \cite{fiat2019decoupling} for ReLU networks. However, decoupling the activation matrix in ReLU networks poses two major problems. First, there is the issue of discontinuity. If the activation pattern jumps when the hidden state is not zero, it introduces a discontinuity in the model. For coupled ReLU, this issue is avoided because the activation pattern only changes when the state is at zero. The second problem is the inability to learn the parameters $\textbf{W}_2$ that generate a good activation pattern. Since there is no gradient flowing through the discrete pattern, it becomes impossible to optimize and update the parameters through gradient-based optimizers. As a result, these parameters can only be randomly initialized, limiting the model's ability to learn an effective activation pattern based on the data.

For IAT-ReLU with continuous input basis $\bp(\cdot)$, we do not have the aforementioned problems. Firstly, the activation pattern is a continuous function of the model input, and after decoupling, the resulting model will still be continuous, which is demonstrated in Figure \ref{fig:actv-pt}. This ensures a continuous mapping between the input and output.  Secondly, for continuous $\bp(\cdot)$, the pattern is determined by the roots and these roots have non-zero gradients flowing through them. This makes active learning of the activation pattern through gradient-based optimizer possible, in contrast to the previous case with ReLU where the pattern has zero gradient everywhere. Thirdly, another advantage is that all activation patterns reside in the same space $[-1,1]$, regardless of the network width. This enables the transfer of patterns between networks with different widths, providing more flexibility. In the case of ReLU, the pattern is an $\bR^d$ vector, and it needs to match the width of another network, limiting its transferability.

\section{Numerical experiments}\label{sec:exp}
\begin{figure}[h!]
  \centering
  {\includegraphics[width=0.9\textwidth]{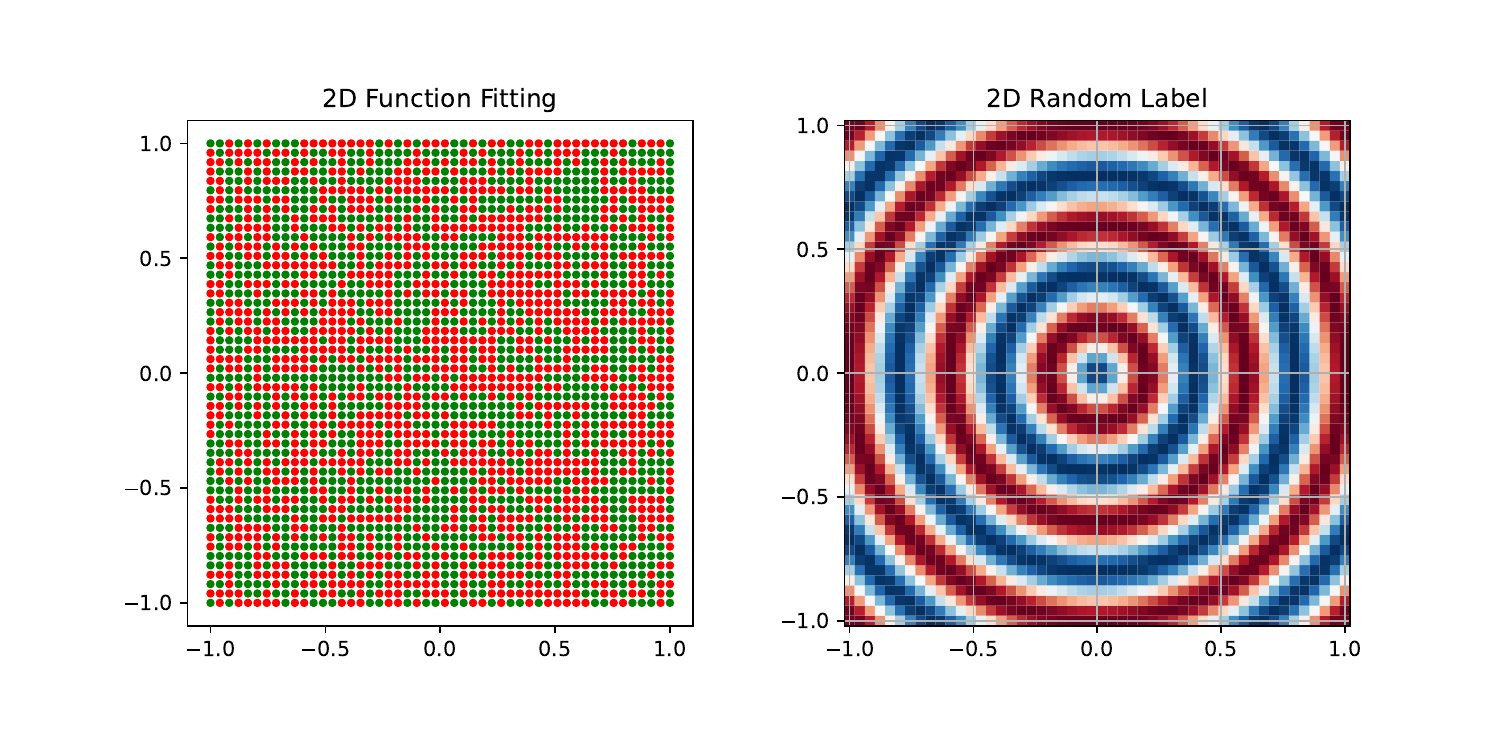}}
  \vspace{-0.4cm}
  \caption[Visualization of the test problems to demonstrate the expressive power.]{Visualization of the test problems to demonstrate the expressive power.  Left: Random label memorization, where color represents the label (-1 and 1) and labels are provided on 50 by 50 uniform mesh points in $[-1,1]^2$. Right: high-frequency function in 2D space. The target function is sampled on 50 by 50 uniform mesh points in $[-1,1]^2$. }
  \label{fig:problem-vis}
\end{figure}

In this section, we conduct numerical experiments to assess the expressive power of IAT-ReLU. We focus on two tasks: fitting a high-frequency function and random label memorization, which are shown in Figure \ref{fig:problem-vis}. Both tasks require a high degree of expressive power to achieve good performance. The first task involves approximating a periodic function using piecewise linear functions, which typically require a large number of linear pieces. The second task evaluates the model's ability to memorize and recall random labels. While smoothness is not crucial for label memorization, the task still demands a high level of expressive power. By examining the performance of IAT-ReLU on these tasks, we can gain insights into its ability to handle complex functions and exhibit exceptional expressive power.

\subsection{Choice of the basis functions}\label{sec:exp-basis}

\begin{figure}[h!]
  \centering
  {\includegraphics[width=0.99\textwidth]{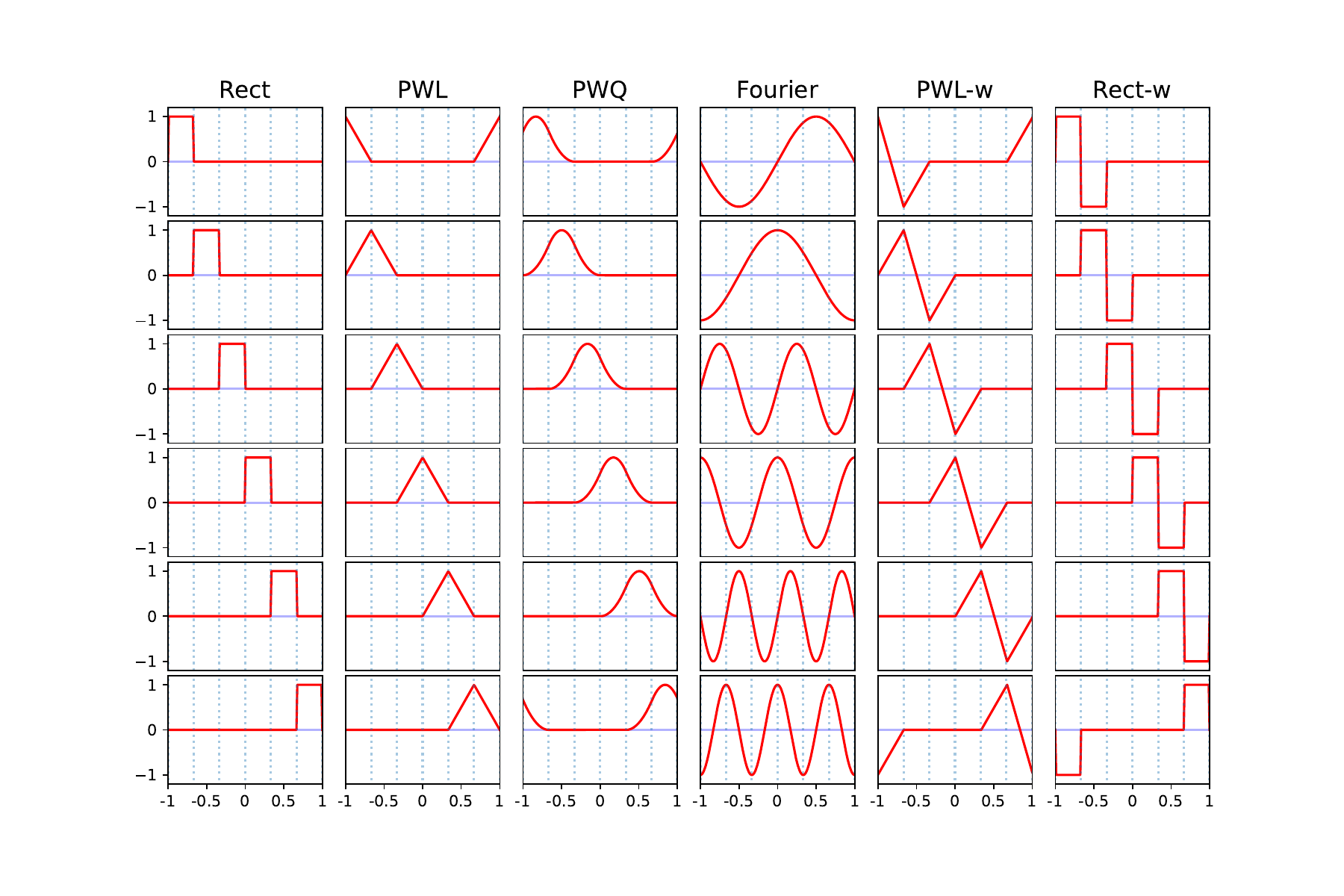}}
  \vspace{-0.8cm}
  \caption{Visualization of the tested basis function with collection size 6. Each column corresponds to one collection of $\bp(\cdot)$. }
  \label{fig:basis-vis}
\end{figure}

When selecting the basis function for IAT-ReLU applied to DNN, we consider four properties: continuity, smoothness, global variation, and zero integral. Based on these properties, six candidate basis functions are included in our numerical experiments as visualized in Figure \ref{fig:basis-vis}, and they are listed as follows. The Rectangular (Rect) basis functions are local, discontinuous, and exhibit a piecewise constant activation pattern.  Choosing the Rect basis as both input and output basis reduces IAT-ReLU to scalar ReLU. The Piecewise Linear (PWL) basis serves as the continuous counterpart to the Rect basis, featuring locally varying activation patterns. Piecewise Quadratic (PWQ) basis functions are smooth versions of Rect basis. The Rectangular Wavelet (Rect-w) basis functions incorporate the zero integral property to enhance the global variation of the activation pattern.  The Piecewise Linear Wavelet (PWL-w) is the continuous version of the Rect-w basis. Lastly, Fourier basis functions are smooth with global variation and zero integral, and their analytic nature ensures a globally continuous activation pattern. 
Notably, although PWL, PWQ, and PWL-w lack globally continuous activation patterns, their tendency for activation pattern jumps is minimal. Therefore, these basis functions can be considered to have an almost continuous activation pattern. The properties of these basis functions are summarized in Table \ref{tab:basis-info}.

\begin{table*}[h!]
    \centering
    \begin{tabular}{|l|cccc|}
     \hline
     & Continuous & Smooth & Global & Zero integral \\
    \hline
    Rect     & \xmark & \xmark & \xmark & \xmark    \\ \hline
    PWL     & \cmark & \xmark & \xmark & \xmark    \\ \hline
    PWQ     & \cmark & \cmark & \xmark & \xmark    \\ \hline
    Fourier & \cmark & \cmark & \cmark & \cmark    \\ \hline
    PWL-w   & \cmark & \xmark & \xmark & \cmark    \\ \hline
    Rect-w  & \xmark & \xmark & \xmark & \cmark    \\ \hline
    \end{tabular}
    \caption{properties for basis }
    \label{tab:basis-info}
\end{table*}

Table \ref{tab:2d-fitting} and Table \ref{tab:2d-mem} present the performance of IAT-ReLU for different combinations of basis functions. Comparing the input basis function, we observe that the basis functions with zero integral (Rect-w, PWL-w, and Fourier basis), outperform the local input basis functions (Rect, PWL, and PWQ). This is because the zero integral basis functions exhibit an enriched activation pattern, leading to increased expressive power and improved trainability. Notably, the Fourier input basis demonstrates the best overall performance across both tasks. Regarding the output basis, the local basis functions (Rect, PWL, and PWQ) exhibit relatively better performance. The best performance in both tasks is achieved with the PWQ output basis. These results demonstrate the effectiveness of combining the input basis functions, based on properties from Section \ref{sec:3.2.2} for promoting the global variation of the activation pattern, with local output basis functions to extract the enriched features.

\begin{table}[h!]
  \centering
  \begin{tabular}{|c |c c c c c c|}
    \multicolumn{7}{c}{IAT-ReLU} \\
    \hline
    \backslashbox[15mm]{$\bp(\cdot)$}{$\bq(\cdot)$} & Rect & PWL & PWQ & Fourier & PWL-w & Rect-w \\
    \hline
        Rect & 89.7\% & 90.5\% & \textbf{92.1}\% & 89.4\% & 86.5\% & 86.1\%  \\ 
        PWL & 85.8\% & 87.3\% & 85.2\% & \textbf{93.0}\% & 86.2\% & 86.5\%  \\ 
        PWQ & 85.5\% & 80.4\% & 80.5\% & 91.6\% & 86.7\% & 88.9\%  \\ 
        Fourier & \textbf{93.2}\% & \textbf{92.5}\% & \textbf{94.1}\% & 89.5\% & 89.5\% & 88.4\%  \\ 
        PWL-w & 90.5\% & \textbf{93.0}\% &\textbf{ 92.3}\% & \textbf{91.9}\% & 82.8\% & 86.3\%  \\ 
        Rect-w & 90.1\% & \textbf{95.1}\% & \textbf{98.3}\% & 89.2\% & 85.7\% & 82.7\%  \\ 
    \hline
  \end{tabular}
  \bigskip\\
  \centering
  \vspace{-1em}
  \begin{tabular}{ c c c}
    \multicolumn{3}{c}{Scalar Activation}\\
    \hline
    ReLU & Tanh & Sigmoid \\
    \hline
    90.4\% & 84.7\% & 83.9\%  \\
    \hline
  \end{tabular}
  \caption{Accuracy for random label memorization task. Top: Performance of IAT-ReLU under different choices of basis function with discretization M=500, where each row corresponds to the same input basis $\bp(\cdot)$ and each column corresponds to the same output basis $\bq(\cdot)$. The best 10 cases are highlighted in boldface.  Bottom: Performance for traditional scalar activation functions.}
  \label{tab:2d-mem}
\end{table}

\begin{table}[h!]
  \centering
  \begin{tabular}{|c |c c c c c c|}
    \multicolumn{7}{c}{Reulst for IAT-ReLU} \\
    \hline
    \backslashbox[15mm]{$\bp(\cdot)$}{$\bq(\cdot)$} & Rect & PWL & PWQ & Fourier & PWL-w & Rect-w \\
    \hline
    Rect & 8.8E-4 & 2.6E-3 & 1.8E-3 & 2.0E-3 & 1.5E-3 & 3.0E-3  \\ 
    PWL & 1.7E-4 & 2.5E-4 & \textbf{7.5E-5} & 3.9E-4 & 2.5E-4 & 1.9E-4  \\ 
    PWQ &\textbf{ 1.6E-4} & \textbf{1.0E-4} & \textbf{6.7E-5} & 5.2E-4 & 2.2E-4 & 2.4E-4  \\ 
    Fourier & \textbf{7.9E-5} & \textbf{7.0E-5} & \textbf{9.9E-5} & \textbf{1.0E-4} & \textbf{7.6E-5} & 1.9E-4  \\ 
    PWL-w & 2.7E-4 & 2.0E-4 & \textbf{1.3E-4} & 2.8E-4 & 3.9E-4 & 3.3E-4  \\ 
    Rect-w & 1.1E-3 & 3.1E-3 & 3.1E-3 & 2.6E-3 & 3.0E-3 & 2.7E-3  \\ 
    \hline
  \end{tabular}
  \bigskip\\
  \centering
  \vspace{-1em}
  \begin{tabular}{ c c c}
    \multicolumn{3}{c}{Scalar Activation}\\
    \hline
    ReLU & Tanh & Sigmoid\\
    \hline
    1.1E-3 & 4.1E-6 & 3.2E-6 \\
    \hline
  \end{tabular}
  \caption{MSE for 2D function fitting problem. Top: Performance of IAT-ReLU under different choices of basis function with discretization M=500, where each row corresponds to the same input basis $\bp(\cdot)$ and each column corresponds to the same output basis $\bq(\cdot)$.The best 10 cases are highlighted in boldface. Bottom: Performance for traditional scalar activation functions.}
  \label{tab:2d-fitting}
\end{table}

\subsection{Choosing discretization M}\label{sec:exp-M}
\begin{figure}[h!]
  \centering
  {\includegraphics[width=0.99\textwidth]{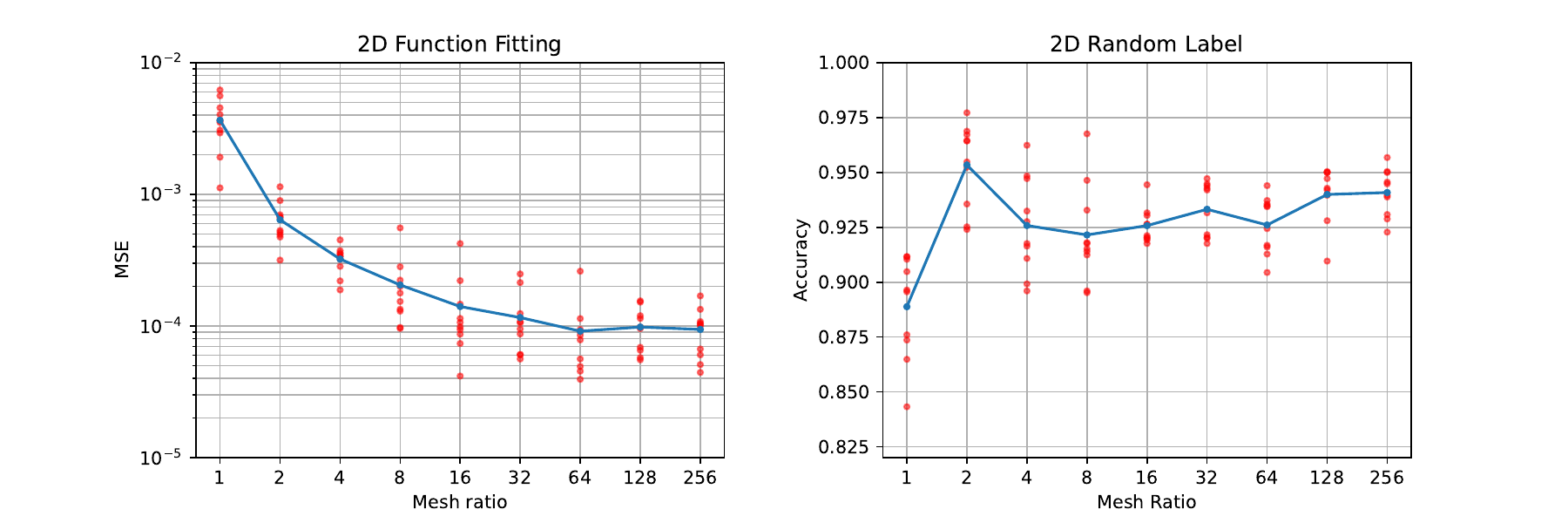}}
  \caption[Performance with respect to discretization.]{Performance with respect to discretization M ($M = d \times \text{mesh ratio}$).  Left: MSE in the function fitting problem with different mesh ratios. Right: Accuracy in label memorization problem with different mesh ratios. }
  \label{fig:change-m}
\end{figure}

Another important aspect of IAT-ReLU is the discretization parameter, M, which controls the smoothness of the solution by determining the number of linear pieces. Figure \ref{fig:change-m} illustrates the results with different values of discretization M, which is parameterized by mesh ratio, i.e., $M = d\times \text{mesh ratio}$. Increasing the discretization offers two key observations. Firstly, for tasks that require smoothness, such as the function fitting task, higher discretization improves the final fitting quality. Conversely, for tasks that do not require smoothness, like the random label memorization task, increasing the discretization yields limited improvements in performance. In other words, the fineness of the linear pieces has minimal impact in such cases. Additionally, finer discretization results in more continuous gradients of the parameters. This is achieved by having more piecewise constant pieces in the solution. Therefore, increasing the discretization enhances training stability, even if the solution does not necessarily demand smoothness.

\section{Conclusion}\label{sec:5}
In this work, we propose the IAT as a means to increase the practical expressive power of DNNs. The IAT utilizes input and output basis functions to perform activation in the function space. 
By choosing different input/output basis functions and nonlinear activations, various versions of IAT can be obtained. In addition, we also showed the equivalence between the IAT and GDNN.
In particular, when the nonlinear activation function is ReLU, we have IAT-ReLU, which serves as a generalization of ReLU. We demonstrate that IAT-ReLU and ReLU share the same forward and backward propagation structure. Moreover, by selecting appropriate basis functions (continuous with large total variation), IAT-ReLU exhibits continuous and enriched activation patterns, enhancing the expressive power of DNNs. Numerical experiments support the superiority of IAT-ReLU over ReLU and other activation functions. This study also suggests several potential avenues for future research.

Firstly, the idea of activation pattern pre-training and transfer can be explored. With the generalized activation pattern proposed in this study, it is possible to transfer a well-trained activation pattern to an untrained model of different sizes. By fixing the transferred activation pattern, the remaining optimization problem becomes simpler, since each pair of weight matrix and bias vector is linear to the model output. Following this direction, Transnet \cite{zhang2023transnet} is introduced, in which the activation pattern of shallow NN is carefully designed and pretrained for transferring. After transferring and fixing the activation pattern, the remaining optimization can be solved by a simple least square. Secondly, training DNNs with decoupled activation patterns can be considered. Instead of fixing the activation pattern, we can train another network solely responsible for providing the activation pattern, while the original network focuses on linear coefficients. This is feasible due to the non-zero gradient flowing through the activation pattern. This approach allows for separate optimization procedures for the activation pattern and linear coefficients. Thirdly, there is room for further optimization of basis functions. In this study, we only scratched the surface by considering six types of basis functions. It is worth exploring and designing additional basis functions to improve performance. In addition to designing new basis functions, we may also consider mixing basis functions of different properties, such as a mixture of local and global basis functions. Lastly, alternative optimization algorithms can be explored. IAT-ReLU represents a new class of activation functions that are 1-homogeneous and smooth when choosing continuous basis functions. Being 1-homogeneous implies that only the angle of the input matters in IAT-ReLU. Therefore, it is possible to investigate other optimization algorithms that focus on controlling the angle of the state vector by updating the parameters. These research directions have the potential to advance our understanding and utilization of IAT-ReLU, paving the way for further enhancements in the expressive power and performance of DNNs.

\bibliographystyle{plain}
\bibliography{reference}

\end{document}